\pdfoutput=1
\documentclass{article}

\usepackage[dblblindworkshop,final]{neurips_2026}
\workshoptitle{Managing Agents that Manage Agents}

\usepackage[T1]{fontenc}
\usepackage[utf8]{inputenc}
\usepackage{amsmath,amssymb}
\usepackage{booktabs}
\usepackage{tabularx}
\usepackage{graphicx}
\usepackage{microtype}
\usepackage[table]{xcolor}
\definecolor{panelblue}{HTML}{EAF2F6}
\definecolor{panelgreen}{HTML}{EDF5F0}
\definecolor{panelamber}{HTML}{FFF5E5}
\usepackage[colorlinks=true,allcolors=blue!55!black]{hyperref}
\hypersetup{pdftitle={UnitBoost: Managing Compound LLM Systems with a Merge Operator, Not a Model},pdfauthor={Xing Zhang, Guanghui Wang, Yanwei Cui, Mengdie Flora Wang, Peiyang He}}
\usepackage[capitalise,noabbrev]{cleveref}
\usepackage{algorithm}
\usepackage{algpseudocode}
\newcolumntype{Y}{>{\raggedright\arraybackslash}X}
\newcolumntype{L}[1]{>{\raggedright\arraybackslash}p{#1}}

\newcommand{\gain}[1]{$+#1$}
\newcommand{\method}{UnitBoost}

\title{\method: Managing Compound LLM Systems\\with a Merge Operator, Not a Model}

\author{%
  Xing Zhang \quad Guanghui Wang \quad Yanwei Cui \quad Mengdie Flora Wang \quad
  Peiyang He\thanks{Corresponding author: \texttt{peiyan@amazon.com}} \\[3pt]
  \normalfont AWS Forward Deployed Engineering
}

\begin{document}

\maketitle

\begin{abstract}
Compound LLM systems often solve a coordination problem by adding a higher-level LLM. The resulting
meta-agent reads workers' outputs, writes the final answer, allocates later calls, and decides when to
stop. It is expressive, but it also concentrates three control decisions in an opaque, order-sensitive
model call. We ask whether the manager needs to be generative at all. \method\ replaces that model with
a defined meta-level operator: a task-given unit map turns worker outputs into slot--value proposals, a
constrained argmax assembles the output, and the slots left unfilled or unsupported become an explicit
residual for the next round. The operator is order-free, records unit provenance, and gives a simple
guarantee: without coupling constraints, unit-wise maximization under the same admission score
dominates selection of any complete candidate. On three held-out benchmarks, it exceeds the best single candidate chosen with gold labels by
0.060--0.195 absolute task-score points and input-matched generative managers by 0.048--0.076. Replacing only the
management step improves six compound-system configurations by 0.013--0.182. Residual-directed rounds
raise FanOutQA cell F1 from 0.4778 to 0.5524; matched controls
show that the true residual outperforms random targets and ordinary rereading, while a label-free
supply signal flags exhaustion after one unproductive round. The same analysis measures three
conditions in which no such gain is available (one indivisible unit, unavailable unit identity,
and an endpoint that charges for every emitted unit) and quantifies cross-unit coupling as a repair
cost. The manager gives up semantic freedom and gains order invariance, unit provenance, and testable
failure conditions.
\end{abstract}

\section{Introduction}

A compound LLM system answers by orchestrating several model calls. Its manager has three jobs: assemble
those workers' partial results into one output, allocate the next calls to whatever remains unresolved,
and decide when further work is no longer useful. Current systems commonly delegate all three jobs to another language model. A
mixture-of-agents aggregator rewrites proposals \citep{moa2024}; debate reports a model-mediated
consensus \citep{debate2023}; iterative refinement asks a model to diagnose and revise its own answer
\citep{madaan2023selfrefine}. Optimizers for language-model programs improve prompts and modules
\citep{dspy2023}, but at inference time the final manager is still usually a generative model.

That choice creates an avoidable control problem. A generative manager can introduce unsupported
content, react to proposal order, and hide which worker supplied which part of its answer. More
fundamentally, recent work finds that judge-based selection outperforms synthesis on its evaluated
tasks \citep{selectionbottleneck2026}, that selecting among models is bounded by the
questions on which they all fail \citep{cofailure2026}, and that full-solution communication can
erase worker diversity \citep{interactiontax2026}. A better selector remains bounded by the best candidate
it receives. It cannot return ``the first part from worker A and the second part from worker B''
unless another model successfully rewrites them.

\begin{figure}[t]
\centering
\includegraphics[width=0.95\linewidth]{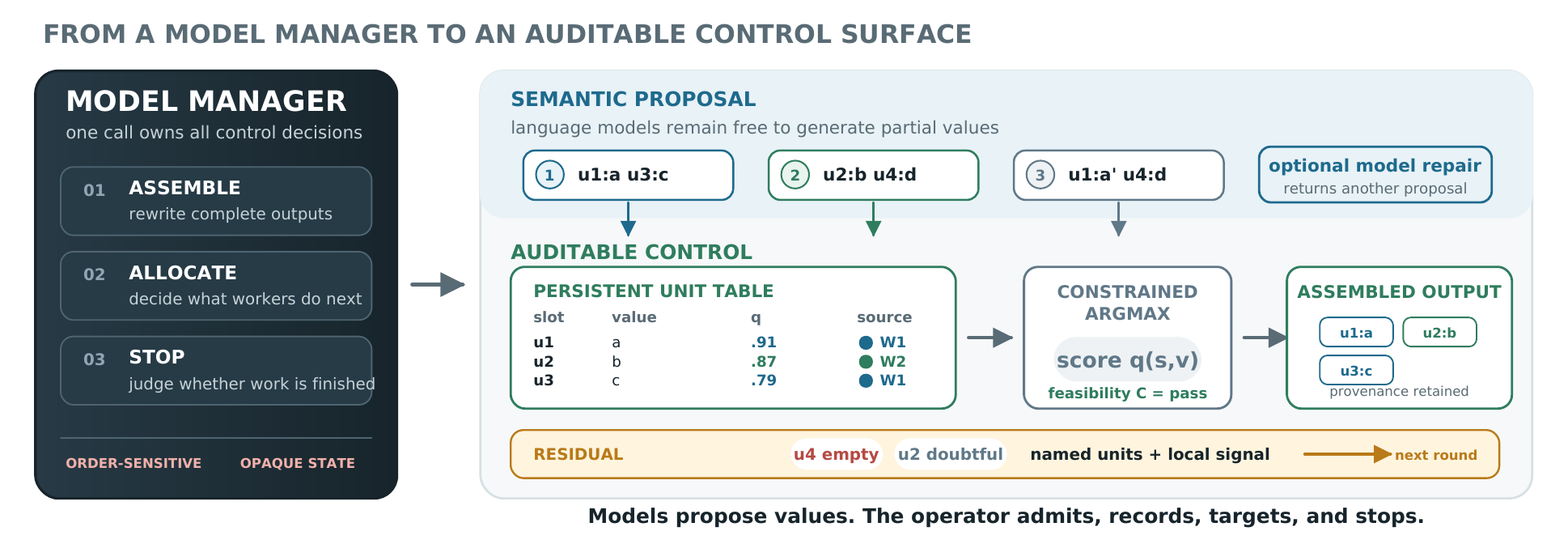}
\caption{Two designs for the same meta-level decisions. A generative manager owns assembly, allocation,
and stopping in one opaque call. \method\ exposes those three decisions as a constrained argmax over a
persistent unit table, a named residual, and a supply-based stop, and keeps the worker and score behind
every accepted value.}
\label{fig:manager}
\end{figure}

We ask whether this manager needs to be a generative agent at all. Our answer is \method: a
deterministic meta-level operator that manages structured parts of outputs rather than complete
outputs. \Cref{fig:manager} contrasts the two control surfaces. The task provides a \emph{unit map}
from an output to proposals $(s,v)$, where slot $s$ identifies a task-defined unit and $v$ is proposed
content for it. For a multi-answer question, a normalized answer identifies a unit; for a table, an
entity identifies a unit and its attribute is the value; for class-level code, a method signature
identifies a unit and its implementation is the value. The operator scores values locally, chooses
across workers at each slot, and enforces a feasibility predicate on the assembled output. Its
residual is then literal: the slots left empty, infeasible, or without an accepted value become the
work queue for the next round.

This change makes a compound system easier to reason about. The merge is a set operation, hence exactly
order-free. Every accepted value has a source, a score, and a feasibility trace. With no cross-unit
constraint, a sum of per-unit maxima is at least the maximum per-candidate sum, so the operator
structurally dominates selection. Most importantly for a multi-round system, outputs from different rounds
enter one persistent unit table. A later worker does not need to rewrite the incumbent; it only needs
to improve one residual unit.

We test the manager as a replaceable system component rather than as a new end-to-end agent. Agents,
prompts, rounds, evidence, and call counts remain fixed while only the candidate-to-output step changes.
The experiments answer four questions:

\begin{enumerate}
\item Can a defined manager pass the ceiling that binds candidate selection?
\item Does it improve existing compound protocols without changing their workers?
\item Does its residual allocate subsequent calls better than additional sampling?
\item Which observable properties predict when the manager will help or fail?
\end{enumerate}

The answer to the first three is yes under one precondition: the task must expose multiple
identifiable units that workers cover differently. The fourth matters as much: the manager cannot help
when the output has a single unit, when equivalent units cannot be identified, or when the endpoint
prices every emitted unit, and feasibility repair taxes whatever gain remains. Each boundary follows
from the operator and is measured rather than merely acknowledged.

\section{Related work}

\paragraph{Generative management.}
Multi-agent frameworks expose programmable conversations and role structures
\citep{autogen2023,metagpt2023}, orchestrators plan and re-plan above specialized workers
\citep{magenticone2024}, a verifier model decides whether a decomposed query needs another round
\citep{verifiedorch2026}, and a meta-agent can invent agent programs outright \citep{adas2024}. In each
design a model owns assembly, allocation, or stopping. \method\ asks which of those decisions still
need model freedom once a worker topology exists, and answers that \emph{models should propose units
while code controls admission, allocation, and stopping}.

\paragraph{Combination and its ceiling.}
Compound systems combine with generative aggregation \citep{moa2024,llmblender2023}, debate
\citep{debate2023}, self-feedback \citep{madaan2023selfrefine}, or module optimization
\citep{dspy2023}, and recent analyses show the limits: candidate selection is the bottleneck
\citep{selectionbottleneck2026}, shared failures bound the benefit of selecting among models \citep{cofailure2026},
and interaction erases useful diversity \citep{interactiontax2026}. Trace-level synthesis
\citep{beyondconsensus2026} and Residual Mixture-of-Agents \citep{rmoa2025} keep a generative combiner.
The closest defined precedent is classical metasearch
\citep{fox1994combination,aslam2001metasearch,cormack2009rrf}, which fuses ranked lists with no model at
all; it is \method's verifier-free special case and a strong baseline here, but it carries no feasibility
predicate, no provenance requirement, and no residual round. Program evolution faces the same choice
between free model-authored edits \citep{alphaevolve2025} and structural crossover with model-based
repair \citep{astcrossover2026}.

\section{The UnitBoost operator}

\paragraph{The management contract.}
A manager should not be evaluated only by the quality of the sentence it returns. It should expose
what it accepted, why each unit won, which constraints were checked, where the next call is spent,
and why the loop stops. \Cref{tab:contract} in \cref{app:detail} states the same four responsibilities
for a generative manager and for \method. The proposal is not to make workers deterministic, but to
keep model freedom out of the control surface that joins proposals into system state.

\paragraph{Interface.}
Let task-defined units be indexed by slots $s$, and let each of the $n$ candidate outputs $y_i$ map
through the unit map $U$ to zero or more proposals $(s,v)$. Write $v_s^i$ for candidate $i$'s value at slot $s$,
using $v_s^i=\bot$ when it proposes none, and let $V_s=\{\bot,v_s^1,\ldots,v_s^n\}$, so leaving a slot
empty is always an option. Let $q(s,v)\geq0$ score a value, with $q(s,\bot)=0$, and let $C$ be a
predicate on assignments $\mathbf{v}=(v_s)_s$, equal to $1$ when the assembled output is feasible. The
manager returns
\begin{equation}
\label{eq:merge}
\operatorname{merge}(y_1,\ldots,y_n)=
\operatorname*{arg\,max}_{\mathbf{v}\,\in\,\prod_s V_s}\sum_s q(s,v_s)
\quad\text{subject to } C(\mathbf{v})=1.
\end{equation}
A canonical normalized-value order breaks ties, and every nonempty choice comes from a worker. If
$C$ is vacuous, the problem separates by slot, and when $C$ factorizes into constraints on disjoint
groups of slots it separates within each group. For coupled code units, we
admit values in score order while running the task's executable check. Across all 95 held-out ClassEval
classes, exhaustive enumeration of candidate method combinations confirms that this order reaches the
product-space optimum. That
agreement is measured rather than guaranteed, and where coupling is dense the same interface accepts an
exact solver over the product of the per-slot value sets, at the cost of enumeration.

\paragraph{Why it can beat every selector.}
With $C$ vacuous,
\begin{equation}
\label{eq:bound}
\sum_s \max_i q(s,v_s^i)\ \geq\ \max_i \sum_s q(s,v_s^i).
\end{equation}
The right side is the best complete candidate under the same score. Equality holds exactly when a single
candidate attains the maximum at every slot. The useful content is therefore not the inequality itself, but its
exception: the manager gains only when workers are incomplete in different places. The output budgets of
the reported testbeds do not break it: with a budget of $B$ values, the top $B$ per-slot maxima score at
least as highly as any candidate's own at-most-$B$ values. A feasibility predicate on content can consume the gain, so the
operative statement is \emph{unit-wise gain minus repair cost}. We measure both rather than assume either.

\paragraph{Three scoring tiers.}
We keep the source of the admission score $q$ explicit; it ranks values and is not a reported task
endpoint. The oracle tier uses reference labels and is a ceiling, never a
method. The deployable tier uses only signals available to the system: worker agreement, the value's
rank in its worker's own list, whether retrieved passages contain it, and a retrieval-backed check generated
and answered by the system. Each coarse signal bin takes its smoothed empirical correct rate from
development questions only. The verifier-free tier uses agreement alone, together with the classical
score and rank fusion rules CombSUM, CombMNZ, Borda, and reciprocal-rank fusion
\citep{fox1994combination,aslam2001metasearch,cormack2009rrf}.

\paragraph{Residual allocation.}
After merging, the manager forms the residual $R_t$, the set of slots that are unfilled or infeasible, or
that hold no accepted positive-score value; we call slots of the third kind \emph{doubtful}. Workers in round $t+1$
receive the unit specification, accepted values, and the local failure signal, but never another
worker's complete solution. New values are inserted into the same table and compete with
incumbents under \cref{eq:merge}. Relative to output $z_{t-1}$, an admission margin $\delta$ makes a
new value at an occupied slot eligible only when its score is at least the incumbent's score plus
$\delta$; values proposed for unfilled slots remain eligible. The argmax is then solved over the
incumbents and the eligible values; in round one, all proposed values are eligible. Call a slot
new in round $t$ when no earlier round proposed a value for it, and let $\rho_t$ be the share of
round-$t$ values that land at new slots; the loop stops once $\rho_t$ falls to a threshold $\tau$.
Each term of the boosting analogy \citep{friedman2001} then has a referent: the merged output is the ensemble, $R_t$
is the residual, workers are weak learners, unit-wise maximization is addition, the margin is
shrinkage, and residual exhaustion is early stopping. Unlike learned boosting, no model parameters are
updated; the object that improves is the compound system's persistent output.
\Cref{alg:unitboost} gives the complete inference loop.

\begin{algorithm}[t]
\caption{\method's inference loop. The unit map, score, feasibility predicate, admission margin, and
stopping threshold are fixed before held-out evaluation.}
\label{alg:unitboost}
\begin{algorithmic}[1]
\Require workers $\mathcal{W}$, unit map $U$, score $q$, predicate $C$, margin $\delta$, rounds $T$,
threshold $\tau$
\State unit table $\mathcal{H}\gets\varnothing$; output $z_0\gets$ empty; residual $R_0\gets$ the whole task request
\For{$t=1,\ldots,T$}
  \State $Y_t\gets\Call{Query}{\mathcal{W},R_{t-1}}$
  \ForAll{$y\in Y_t$}
    \State insert $(s,v,\text{worker},t)$ from $U(y)$ into $\mathcal{H}$
  \EndFor
  \State $z_t\gets\Call{ConstrainedUnitArgmax}{\mathcal{H},q,C,z_{t-1},\delta}$
  \State $R_t\gets\{s:s\text{ is unfilled, infeasible, or has no accepted positive-score value}\}$
  \State $\rho_t\gets$ share of round-$t$ values that land at new slots
  \If{$R_t=\varnothing$ \textbf{or} $\rho_t\leq\tau$}
    \State \textbf{break}
  \EndIf
\EndFor
\State \Return $z_t$ and the source, score, and constraint trace of every accepted value
\end{algorithmic}
\end{algorithm}

\section{Experimental setup}

\paragraph{Testbeds.}
QAMPARI asks questions with many entity answers distributed across passages
\citep{qampari2022}; we use 200 development and 800 held-out questions. ASQA represents an ambiguous
question by a set of distinct readings \citep{asqa2022}; we use 200 and 748. Both use the ALCE retrieval
corpus and prompts \citep{alce2023}. FanOutQA pairs questions with human-written decompositions over
Wikipedia articles \citep{fanoutqa2024}; we score answers that form an entity--attribute table, with one
article per entity. 85 development and 179 held-out questions have complete evidence. ELI5
\citep{eli52019} tests long-form factual statements; we use 200 and 800. Archived SWE-bench
\citep{swebench2023} and ClassEval
\citep{classeval2023} outputs probe single-unit and coupled code. \Cref{tab:testbeds} in
\cref{app:detail} summarizes their unit interfaces, identity sources, and splits.

\paragraph{Evaluation endpoints and statistics.}
Task scores are external to $q$, macro-averaged over questions on $[0,1]$. QAMPARI uses capped recall
$\min\{H/\min(G,5),1\}$ under a 20-answer budget ($H$ distinct reference hits, $G$ references);
ASQA uses reading coverage under 12 answers; ELI5, atomic-claim coverage under five statements; and
FanOutQA, answer-table cell F1. ClassEval reports how often a method that passes its own tests
still passes in the assembled class. Below their caps the first three endpoints are additive in
units found, which is the regime \cref{eq:bound} describes; set F1 is a separate boundary.
Held-out differences use 4,000 one-sided paired bootstrap resamples; the residual-size correlation uses
4,000 permutations. All choices use development data; $p$-values are uncorrected.

\paragraph{Workers and managers.}
The primary retrieval system has ten workers, each reading a disjoint window of ten passages from the
same top-100 list. The controls are one call reading all passages, ten stochastic rereads of that call,
ten independent workers sharing the same evidence, a judge model, generative managers with and without
the evidence, answer-level voting, rank fusion, and oracle candidate selection, which uses gold labels to
pick the single best complete candidate and is shortened to oracle selection below. DeepSeek-V3.2
\citep{deepseekv32} supplies the primary workers and matched manager, while Qwen3-32B \citep{qwen3}
supplies the held-out worker replication. Claude Sonnet 5 \citep{claudesonnet5} and GPT-5.6 Sol \citep{gpt56}
manage the same DeepSeek proposals; Claude Sonnet 5 also supplies the exploratory development pool.
\Cref{tab:modelroles} lists the worker and manager roles.

\section{Passing the candidate-selection ceiling}

\begin{table}[t]
\centering
\setlength{\tabcolsep}{3.8pt}
\renewcommand{\arraystretch}{1.08}
\caption{\textbf{Held-out manager replacement.} Workers, prompts, evidence, and calls are fixed; only
candidate-to-output management changes. Panel A replaces one generation-round manager, panel B one
protocol step. Endpoints are QAMPARI capped recall/20 answers, ASQA coverage/12 answers, and FanOutQA
cell F1. Each $p$ is the fraction of paired bootstrap resamples with non-positive gain.}
\label{tab:main}
\begin{tabularx}{\linewidth}{@{}p{0.24\linewidth}Yrrrr@{}}
\toprule
\textbf{setting} & \textbf{replaced manager} & \textbf{comparator} & \textbf{\method} &
\textbf{gain} & \textbf{$p$} \\
\midrule
\rowcolor{panelblue}\multicolumn{6}{@{}l}{\textbf{A. Passing the candidate-selection ceiling}} \\
QAMPARI / 20 answers & oracle candidate selection & 0.3967 & \textbf{0.4685} &
\gain{0.0718} & $<.001$ \\
ASQA / 12 answers & oracle candidate selection & 0.3595 & \textbf{0.4191} &
\gain{0.0596} & $<.001$ \\
FanOutQA / cell F1 & oracle candidate selection & 0.2829 & \textbf{0.4778} &
\gain{0.1949} & $<.001$ \\
QAMPARI / 20 answers & evidence-reading synthesis & 0.4205 & \textbf{0.4685} &
\gain{0.0480} & $<.001$ \\
ASQA / 12 answers & evidence-reading synthesis & 0.3427 & \textbf{0.4191} &
\gain{0.0764} & $<.001$ \\
\midrule
\rowcolor{panelgreen}\multicolumn{6}{@{}l}{\textbf{B. Drop-in management inside compound protocols}} \\
QAMPARI / mixture & generative aggregation & 0.4170 & \textbf{0.4627} &
\gain{0.0457} & $<.001$ \\
ASQA / mixture & generative aggregation & 0.3472 & \textbf{0.4117} &
\gain{0.0645} & $<.001$ \\
QAMPARI / debate & generative consensus & 0.2913 & \textbf{0.4148} &
\gain{0.1235} & $<.001$ \\
QAMPARI / critic & generative revision & 0.2735 & \textbf{0.4258} &
\gain{0.1522} & $<.001$ \\
QAMPARI / chain & final agent's output & 0.2415 & \textbf{0.4235} &
\gain{0.1820} & $<.001$ \\
ASQA / roles & generative integration & 0.3295 & \textbf{0.3424} &
\gain{0.0129} & $.0280$ \\
\bottomrule
\end{tabularx}
\end{table}

\begin{figure}[t]
\centering
\includegraphics[width=0.95\linewidth]{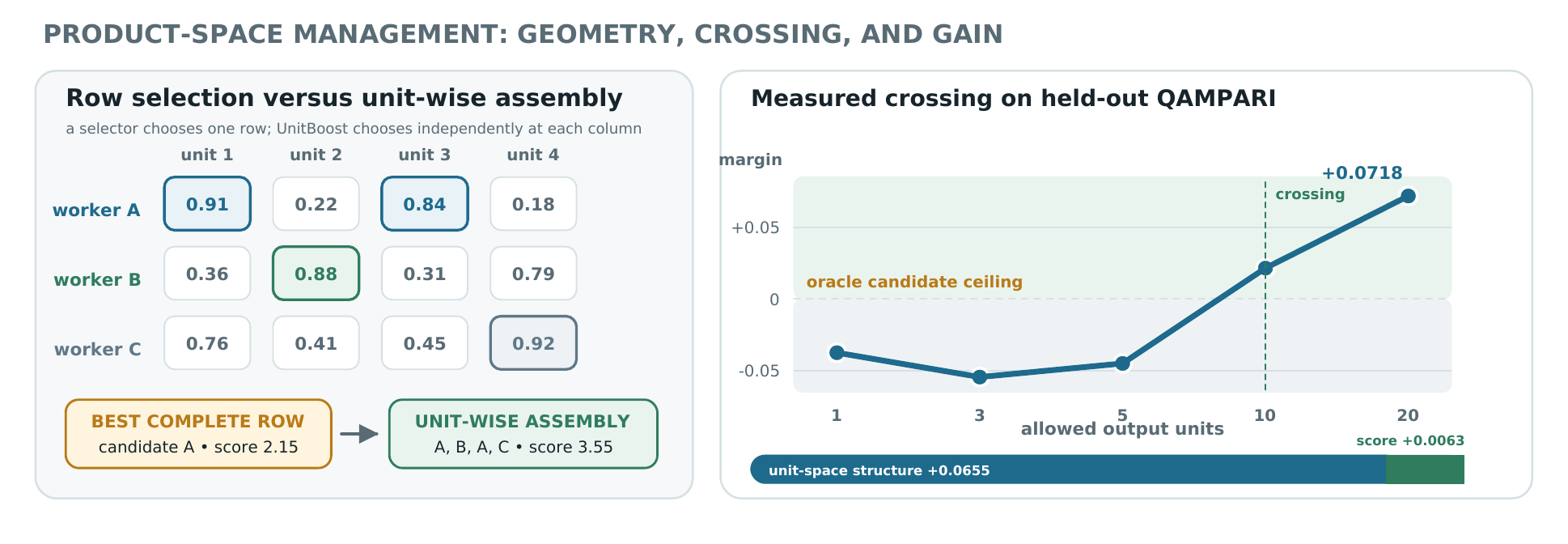}
\caption{Candidate selection is restricted to one row of the score matrix, while unit-wise assembly
searches the product space of its columns. On QAMPARI, the margin over the candidate an oracle would
select turns positive as the output exposes more units. The bar splits that margin into unit-space
structure, which needs no labels, and \method's score.}
\label{fig:margin}
\end{figure}

Panel A of \cref{tab:main} isolates the management step. On QAMPARI, \method\ reaches 0.4685 against
0.3967 for the candidate an oracle would select, \gain{0.0718} at $p<0.001$. On ASQA the margin is
\gain{0.0596}. At one allowed unit the two spaces coincide and \method\ does not win; the margin
crosses zero as the output budget grows, that is, as the number of independently choosable units
increases (\cref{fig:margin}). FanOutQA provides a stricter slot--value test: choosing one article-level
worker reaches only 0.2829, while unit-wise management reaches 0.4778, a margin of \gain{0.1949}.
This benchmark is also a useful boundary: a generative manager that rewrites the proposals reaches
0.6019, but it reads every retrieved article, which the merge never does, so the comparison is not
input-matched. FanOutQA therefore supports the product-space and allocation claims,
not universal dominance over generative managers.

\paragraph{The manager, not the score, supplies most of the gain.}
On QAMPARI, worker order reaches 0.4592, agreement ranking 0.4622, classical fusion
0.4540--0.4630, and the deployable admission rule 0.4685. Reciprocal-rank fusion already exceeds oracle
candidate selection by 0.0655 without labels. \method's scoring adds 0.0063 over that rule and 0.0055
over the best label-free rule tested; choosing in unit space adds the rest. A label-budget sweep over frozen outputs agrees: ten labeled questions already retain
0.0618 of the final 0.0718 margin over oracle selection, and the deployable score first exceeds the best
label-free rule at 25 labels for a 20-unit budget, 50 for tighter budgets.

\paragraph{Can frontier models propose without controlling assembly?}
Claude Sonnet 5 and GPT-5.6 Sol receive the same DeepSeek-V3.2 proposals and are explicitly told the
output budget. Claude Sonnet 5 reaches 0.4918 when unrestricted, above \method's 0.4685, but does so by
emitting units that no worker proposed. After filtering each generated answer to the candidate union, every
configuration is below \method\ at $p\leq0.0005$. Treating Claude Sonnet 5's unrestricted output as an
eleventh proposal instead raises the controlled merge to 0.5132, so a generative manager can supply
novel units without controlling final assembly.

\paragraph{The result survives a different worker model.}
Replacing DeepSeek-V3.2 with Qwen3-32B on the same 800 held-out QAMPARI questions preserves the
crossing budget: \method\ reaches 0.4180, exceeds oracle candidate selection by 0.0615, and exceeds
the matched Qwen3-32B evidence-reading manager by 0.0398. Separately, with Claude Sonnet 5 supplying
all ten window workers on the 200 development questions, the margin over oracle candidate selection is
0.0700. We report this last result as exploratory because the Claude Sonnet 5 pool was not generated
on held-out questions.

\paragraph{Replacing management inside existing systems.}
The operator is useful only if it survives the protocol around it. We therefore replace one step in
five compound protocol families in six configurations, keeping their workers, prompts, evidence,
rounds, and calls unchanged. Panel B of \cref{tab:main} compares \method\ with each protocol's own
manager. It improves every configuration, from
0.0129 on the authored-role pipeline to 0.1820 on the sequential chain, and pooling the units of every debate round rather than only the
last gains a further 0.0467, because the persistent table does not discard earlier useful units.
\Cref{tab:protocols} defines the six configurations and the management step replaced in each.

The order-free property is exact for a fixed candidate set. By contrast, permuting the same five
workers in a sequential chain changes its output score on 57.5\% of questions. Exchange also changes
what management should prefer. With shared evidence, debate raises worker overlap from 0.520 to 0.790;
with partitioned evidence, it moves only 0.233 to 0.256. A protocol that reports one worker's output is
therefore best served by full-solution exchange, which raises the best candidate, while \method\ gains
0.0150 from critique-only exchange, which preserves complementary units. Whether diversity is waste or
supply is therefore a property of the manager, not of the exchange.

\section{Residual-directed allocation and stopping}

\begin{figure}[t]
\centering
\includegraphics[width=0.95\linewidth]{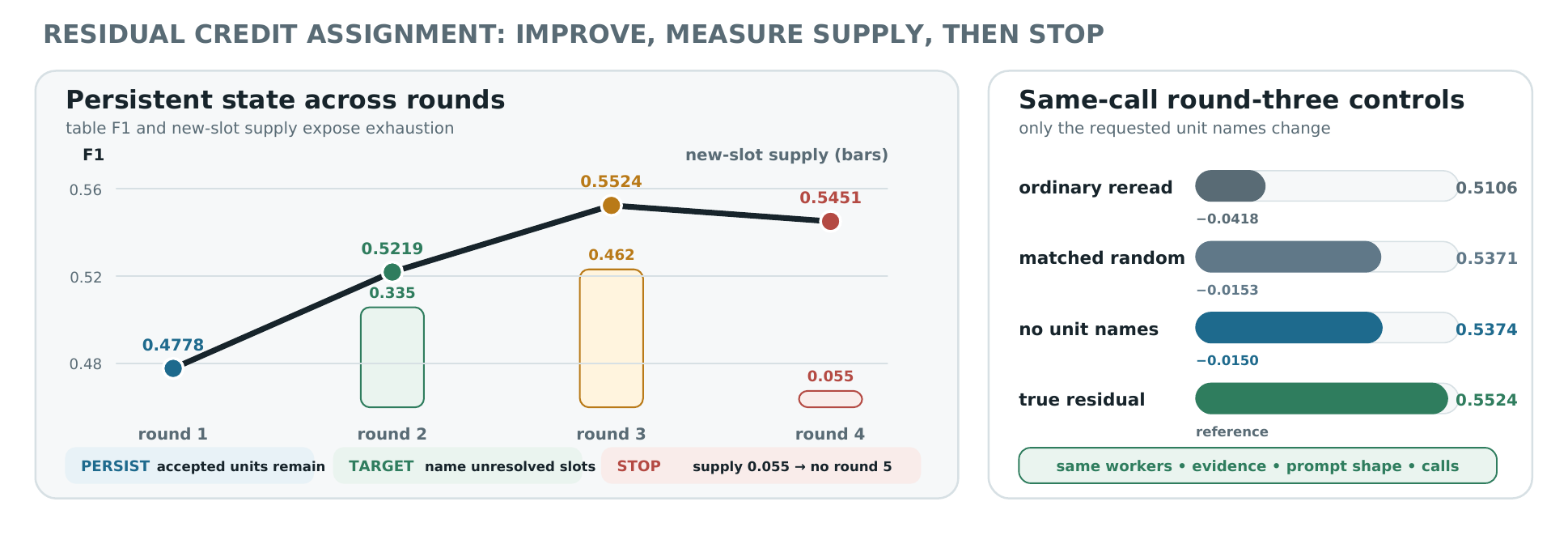}
\caption{Residual credit assignment on FanOutQA. Accepted units persist across rounds; the right panel
holds calls fixed across allocation controls.}
\label{fig:residual}
\end{figure}

The residual turns credit assignment into a set operation. A unit that is already filled and supported
receives no further call; an unfilled, infeasible, or doubtful unit becomes eligible for the next
worker budget. This is deliberately narrower than asking a coordinator to write a global critique:
the manager identifies \emph{where} work is needed, while workers retain responsibility for
\emph{what} content to propose. \Cref{fig:residual} shows the persistent state and matched allocation
controls, while \cref{tab:diagnostics} reports the corresponding held-out tests and boundary
measurements.

A second round that pays does not by itself establish useful management: any extra sampling may
discover new values. FanOutQA makes the control precise because the benchmark names each slot. After
two rounds, we spend the third round's calls four ways, with identical workers, articles, calls,
temperature, accepted and doubtful values, and prompt template. Only the requested slot names differ:
the true residual, a matched random draw from the same unit map, no names, or an ordinary rereading.

\begin{table}[t]
\centering
\setlength{\tabcolsep}{3.5pt}
\renewcommand{\arraystretch}{1.07}
\caption{\textbf{Held-out credit-assignment and boundary diagnostics.} Each row reports the difference
between the two arms it names, in that order, and $p$ is the fraction of paired bootstrap resamples in
which this difference is non-positive. Panel B holds round-three workers, evidence, and calls fixed. The
last two rows of panel C are therefore costs the operator pays, not gains.}
\label{tab:diagnostics}
\begin{tabularx}{\linewidth}{@{}p{0.16\linewidth}Y
  >{\raggedleft\arraybackslash}p{0.21\linewidth}
  >{\raggedleft\arraybackslash}p{0.16\linewidth}r@{}}
\toprule
\textbf{diagnostic} & \textbf{comparison} & \textbf{outcome} & \textbf{delta} & \textbf{$p$} \\
\midrule
\rowcolor{panelblue}\multicolumn{5}{@{}l}{\textbf{A. Persistent residual loop on FanOutQA (cell F1)}} \\
round 2 & against round 1 & 0.5219 & \gain{0.0441} & $<.001$ \\
round 3 & against rounds 1--2 & \textbf{0.5524} & \gain{0.0306} & $.0023$ \\
round 4 & against round 3, after collapse & 0.5451 & $-0.0073$ & $.8715$ \\
\midrule
\rowcolor{panelgreen}\multicolumn{5}{@{}l}{\textbf{B. Same-call round-three allocation}} \\
true residual & random residual & 0.5524 vs.\ 0.5371 & \gain{0.0153} & $.0382$ \\
true residual & untargeted & 0.5524 vs.\ 0.5374 & \gain{0.0150} & $.0755$ \\
true residual & rereading & 0.5524 vs.\ 0.5106 & \gain{0.0418} & $<.001$ \\
\midrule
\rowcolor{panelamber}\multicolumn{5}{@{}l}{\textbf{C. Measured preconditions and costs}} \\
divisibility & full-context pool vs.\ oracle selection & never crosses & $<0$ throughout & -- \\
unit identity & embedding vs.\ lexical units (ELI5) & 0.5092 vs.\ 0.5067 & \gain{0.0025} & $.3698$ \\
oracle identity & reference vs.\ lexical units (ELI5) & 0.6479 vs.\ 0.5067 & \gain{0.1413} & $<.001$ \\
coupling cost & standalone vs.\ sibling-calling methods & 0.985 vs.\ 0.947 & \gain{0.0379} & $.0455$ \\
endpoint cost & oracle selection vs.\ \method, set F1 & -- & \gain{0.0650} & $<.001$ \\
\bottomrule
\end{tabularx}
\end{table}

Panel B of \cref{tab:diagnostics} reports the outcome. The true residual beats the matched random draw
and the ordinary rereading; its advantage over the untargeted prompt is not significant on its own, and
the stronger evidence is a dose response. Relative to the untargeted prompt, naming one residual
slot is worth $-0.0092$, naming two \gain{0.0263}, and naming three or more \gain{0.0516}; the rank
correlation between residual size and gain is $0.183$ at $p=0.0075$.

Rounds accumulate because all values remain in one table, and the label-free supply signal stays high
while the rounds pay and then collapses (\cref{fig:residual}): $\rho_t$ reaches 0.055 in round four, the
round that loses 0.0073. The same
pattern holds on QAMPARI, where the two-signal arm
without the retrieval-backed check improves from 0.4657 to 0.4843 in round two, while round three adds
nothing and new answers per question fall from 5.32 to 2.34. The deployable three-signal arm in
\cref{tab:main} starts at 0.4685. The signal is therefore one round late: it halts the loop after the
unproductive round rather than before it. No held-out score is selected by oracle stopping: the
three-round budget was fixed on the development split, where the fourth round already fell to a supply of
0.0945 and lost 0.0022.

\paragraph{Shrinkage is secondary to coverage.}
The admission margin $\delta$ of \cref{alg:unitboost} is the direct analogue of a boosting learning rate.
Sweeping it from 0 to 0.40 changes held-out QAMPARI by at most 0.0023; on ASQA, a margin of 0.10 adds
0.0036. The large round gains instead come from values at units no earlier worker filled, so the
stagewise component that matters here is persistent coverage expansion rather than shrinkage.

The loop adds more than reordering. On ASQA it exceeds an oracle ordering of every unit available in
round one by 0.0145; on FanOutQA, round two exceeds the previous table's oracle ordering by 0.0237. No
admission rule over earlier values can create a slot value no worker proposed: residual allocation
expands the table rather than rescoring it.

\section{Preconditions and boundaries}

\begin{figure}[t]
\centering
\includegraphics[width=0.95\linewidth]{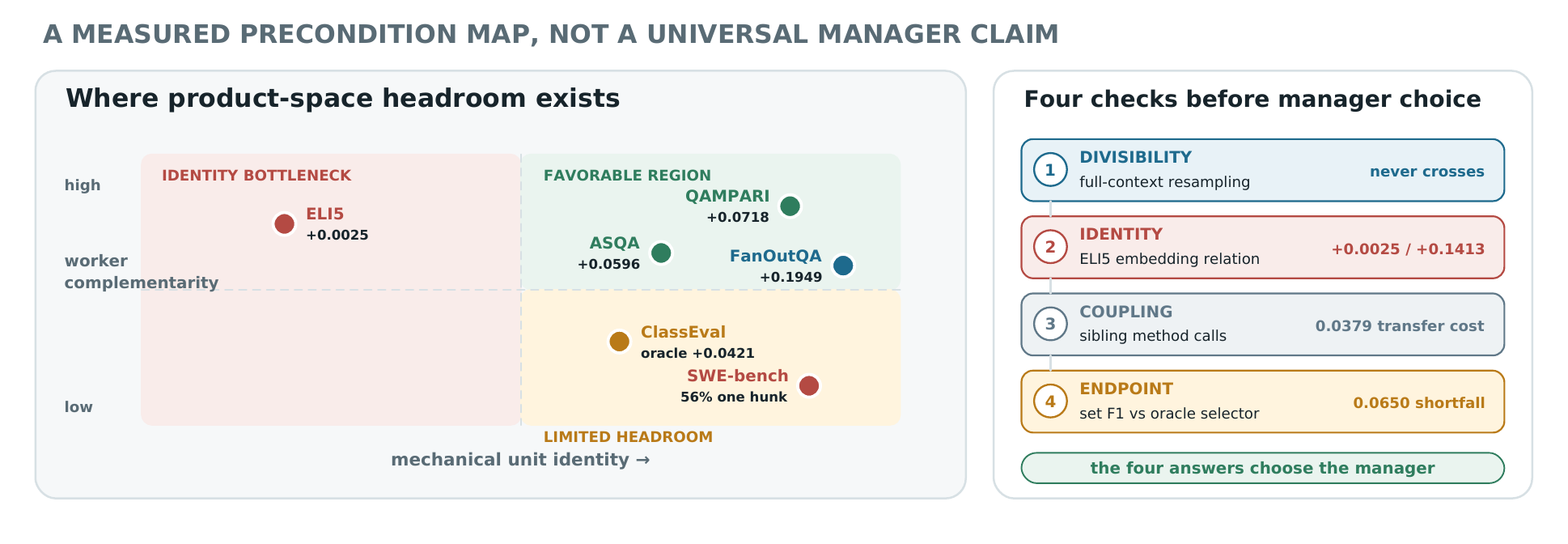}
\caption{Measured operating map. Unit identity and complementarity create product-space headroom;
coupling and the endpoint determine whether it survives. ClassEval is an oracle ceiling.}
\label{fig:boundary}
\end{figure}

\Cref{fig:boundary} locates each testbed by unit identity and worker complementarity, then records the
coupling and endpoint checks that determine whether product-space headroom survives; panel C of
\cref{tab:diagnostics} gives the held-out test behind each. Together they form a pre-deployment
checklist: does the task expose more than one unit, can equivalent units be identified mechanically, do
workers fail on different units, and can locally selected values be admitted without excessive repair or
emission cost?

\paragraph{Complementarity must be manufactured.}
Ten workers reading disjoint evidence produce individually incomplete outputs; ten stochastic samples of
one worker reading all evidence do not. Holding calls and the available passage set fixed, \method\ never
passes oracle selection on the resampled pool at any output budget, while its margin on partitioned
evidence grows with worker count. Partitioning therefore buys the crossing of the oracle ceiling, not the
gain over a deployable manager: on the resampled pool \method\ still scores 0.4878 against 0.4550 for the
candidate the deployable score prefers, and replacing a debate's consensus step is worth \gain{0.0268} at
$p<0.001$ under shared evidence. Better retrieval increases rather than removes the effect: weak,
standard, and oracle-reranked evidence yield margins of 0.0522, 0.0718, and 0.1435 over oracle
selection. What the manager needs is divisible work, not weak evidence.

\paragraph{Unit identity is load-bearing.}
Long-form factual generation has many additive statements but no reliable mechanical identity between
paraphrases. A lexical relation catches only 5.9\% of same-fact pairs at a 1.3\% false-merge rate.
Sentence embeddings improve pair discrimination from 0.635 to 0.672, yet add only 0.0025 to
five-statement atomic-claim coverage. Perfect reference identity would add 0.1413 and reverse the
comparison with the generative manager.
The failure is therefore not a badly tuned threshold: most of the missing gain lies in knowing that two
differently worded statements occupy the same semantic unit.

\paragraph{Coupling appears as repair cost.}
On class-level code, a value from a method that passes its own tests transfers into the assembled class
0.985 of the time when the method stands alone and 0.947 when it calls siblings. The 0.038 coupling
gap is nominally significant at $p=0.0455$. Requiring the transferred method and all siblings to remain clean
gives 0.963 versus 0.920, a 0.043 gap at $p=0.0850$. Exhaustive product search and score-order
admission agree on all 95 held-out ClassEval classes. Under a gold per-unit score, the resulting assembly ceiling is
0.0421 above oracle candidate selection ($p=0.126$); this is an oracle analysis, not a deployable
result. On SWE-bench, 280 of 500 reference patches contain one hunk, so the product space often
collapses to candidate space before repair is considered.

\paragraph{The endpoint sets a price.}
Set F1 charges every answer through precision. Under that endpoint, \method\ beats every deployable manager tested but falls 0.0650 below oracle
selection. A price rule predicts the reversal: combine only while the next unit's precision exceeds what
the metric charges for it. It gets 36 of 40 combine-versus-select decisions right across candidate pools
and selectors, against 24 for a constant decision. Management should be chosen from the task's
unit structure and endpoint, not installed by default.

\section{Discussion}

A meta-agent need not be a language model end to end. \method\ leaves semantic proposal and repair to
models while making admission, allocation, and stopping explicit and auditable. \Cref{tab:lessons}
pairs each design rule this licenses with the held-out measurement behind it.

\paragraph{The gain is control, not free compute.}
The merge itself makes no model call, but deployable scoring adds one retrieval-backed check per value
and complementary workers cost calls; \cref{app:detail} gives the per-question token counts. \method\
extracts more value from fixed workers and allocates later rounds; it does not make workers cheaper.

\section{Limitations}

Calibration is benchmark-specific; transfer without refitting remains untested. Agreement and rank
fusion are label-free alternatives. The score lacks per-worker trust and detects
residual exhaustion one round late. Positive results come mostly from knowledge-intensive language
tasks; the code testbeds measure coupling. Auditable management can still encode a bad score or
feasibility predicate.

\section{Conclusion}

\method\ replaces generative management with a constrained unit-wise argmax over a persistent table
and an explicit residual. When workers cover identifiable units differently, it crosses the
candidate-selection ceiling while preserving provenance, and it states the conditions under which it
cannot. The question this raises for meta-agent design is not whether a manager should be generative,
but which of its decisions still require a model.

\section*{Responsible-use statement}

\method's source records help audit decisions but do not establish factual support. Shared errors
can receive high agreement scores, calibration may fail on new corpora, and residual requests can
assemble sensitive personal profiles from scattered documents. Composed code can violate assumptions
beyond available tests. Deployments should restrict evidence access and unauthorized profiling,
verify consequential claims, recalibrate on target-domain data, cap calls, and require full-suite
testing and human review before code execution. Dividing evidence does not guarantee independent
errors; adversarial robustness is untested. We collect no new human-participant data, but public
benchmarks can contain identifying or objectionable text. Token accounting measures neither energy
use nor environmental impact.

\bibliographystyle{plainnat}
\bibliography{refs}

\appendix

\section{Additional experimental detail}
\label{app:detail}

This appendix collects five reference tables cited from the main text, then gives details behind three
claims. \Cref{tab:contract} states the management contract of \cref{eq:merge};
\cref{tab:testbeds,tab:modelroles,tab:protocols} list the testbeds, model roles, and compound
protocols; and \cref{tab:lessons} pairs each design rule with its measurement.

\begin{table}[ht]
\centering
\caption{Generative and operator implementations of the same management responsibilities.}
\label{tab:contract}
\begin{tabular}{@{}lll@{}}
\toprule
responsibility & generative manager & \method \\
\midrule
assemble & rewrite complete proposals & select proposed values per unit \\
allocate & describe what seems missing & return named residual units \\
stop & self-judge whether done & monitor new-unit supply \\
audit & optional natural-language rationale & source, score, and constraint trace \\
\bottomrule
\end{tabular}
\end{table}

\begin{table}[ht]
\centering
\caption{Testbed unit interfaces and development/test splits.}
\label{tab:testbeds}
\begin{tabular}{@{}llllr@{}}
\toprule
testbed & task-defined unit & identity source & coupling check & dev/test \\
\midrule
QAMPARI & entity answer & normalized string & none & 200/800 \\
ASQA & question reading & accepted aliases & none & 200/748 \\
FanOutQA & entity attribute & benchmark key & one value per slot & 85/179 \\
ELI5 & atomic fact & lexical/embedding map & none & 200/800 \\
ClassEval & method body & method signature & executable class tests & --/95 \\
SWE-bench & patch hunk & diff parser & apply and test & --/500 \\
\bottomrule
\end{tabular}
\end{table}

\begin{table}[ht]
\centering
\small
\caption{Model roles in the reported language-task experiments. The retrieval-backed per-value check and
the judge-model baseline always run on the same model as the candidate pool they score, so they add no
model beyond this list.}
\label{tab:modelroles}
\begin{tabularx}{\linewidth}{@{}L{0.19\linewidth}L{0.19\linewidth}L{0.20\linewidth}Y@{}}
\toprule
role & model & data & use \\
\midrule
primary worker and matched manager & DeepSeek-V3.2 &
QAMPARI, ASQA, FanOutQA, ELI5 & main candidate pools and generative-manager baselines \\
worker-model replication & Qwen3-32B &
QAMPARI held-out (800) & replaces all ten workers in the held-out robustness test \\
frontier manager & Claude Sonnet 5 &
QAMPARI held-out (800) & unrestricted and candidate-restricted manager; unrestricted output is also
admitted as an eleventh proposal \\
frontier manager & GPT-5.6 Sol &
QAMPARI held-out (800) & same manager-only control over the DeepSeek-V3.2 proposals \\
frontier proposer & Claude Sonnet 5 &
QAMPARI development (200) & replaces all ten workers; reported as exploratory \\
\bottomrule
\end{tabularx}
\end{table}

\paragraph{Frontier-manager controls.}
Claude Sonnet 5 and GPT-5.6 Sol each receive the same ten DeepSeek-V3.2 proposals, the same prompt
shape, and an explicit answer-count budget. We evaluate each generated answer before and after
filtering out values absent from the worker pool. This is post-generation filtering, not constrained
generation: the unrestricted answer may contain values from the manager's own knowledge. Both
post-filtered managers fall below \method. When the unrestricted Claude Sonnet 5 list is admitted
as an eleventh proposal, every surviving unit retains its source, score, and constraint trace.

\begin{table}[ht]
\centering
\small
\caption{Compound-protocol configurations used for the drop-in manager comparison. Every configuration
keeps its own workers, prompts, evidence, rounds, and call count; only the replaced step changes.}
\label{tab:protocols}
\begin{tabularx}{\linewidth}{@{}L{0.16\linewidth}L{0.11\linewidth}Y
  L{0.22\linewidth}L{0.17\linewidth}@{}}
\toprule
protocol & testbed & worker interaction & replaced step & endpoint \\
\midrule
mixture of agents & QAMPARI & two layers; layer two sees every layer-one list &
final generative aggregator & capped recall / 20 answers \\
mixture of agents & ASQA & two layers; layer two sees every layer-one list &
final generative aggregator & reading coverage / 12 answers \\
debate & QAMPARI & five agents exchange complete lists for three rounds &
final consensus & capped recall / 20 answers \\
critic--refine & QAMPARI & critics send comments; each author revises its own list &
final revision & capped recall / 20 answers \\
sequential chain & QAMPARI & five ordered agents repeatedly update one list &
last agent's list & capped recall / 20 answers \\
authored roles & ASQA & a coordinator assigns five aspects; workers feed an integrator &
final integrator & reading coverage / 12 answers \\
\bottomrule
\end{tabularx}
\end{table}

\paragraph{Feasibility and provenance.}
Every emitted unit retains the worker that supplied it, its local score, and the checks evaluated before
admission. On FanOutQA, agreement ranking reaches 0.4848 versus 0.4778 for the calibrated ranking; signal
selection on 85 development questions does not transfer, but all seven tested signal subsets remain
more than 0.19 above oracle candidate selection. The structural claim is insensitive to the scorer.

\begin{table}[ht]
\centering
\caption{Design rules supported by manager replacements at fixed worker calls. Each rule is stated only
where a held-out measurement in the body supports it.}
\label{tab:lessons}
\begin{tabularx}{\linewidth}{@{}p{0.15\linewidth}Yp{0.29\linewidth}@{}}
\toprule
decision & measured evidence & design rule \\
\midrule
manager role & a frontier manager helps most when admitted as another proposal &
models propose; admission stays explicit \\
decision space & the unit product beats oracle selection by 0.0596--0.1949 &
use the task-valid unit space \\
call allocation & the true residual beats random targets by 0.0153 at equal cost &
target named residual units \\
stopping & new-slot supply falls to 0.055 in the 0.0073-loss round &
stop when supply collapses (one-round lag) \\
architecture & identity, complementarity, coupling, and endpoint predict outcomes &
choose from task structure and endpoint \\
\bottomrule
\end{tabularx}
\end{table}

\paragraph{Token accounting.}
On the primary retrieval testbed, one worker reading one ten-passage window consumes 5,108 input tokens
per question, one worker reading all passages 18,879, the ten-worker partition 51,128, and ten full-list
resamples 188,788. The deterministic merge makes no model call; the deployable per-value check does.
A residual round costs the same calls as its worker round; its benefit is allocation, not lower
per-round cost.

\end{document}